# LatentBurst: A Fast and Efficient Multi Frame Super-Resolution for Hexadeca-Bayer Pattern CIS images

Sangwook Baek    Vinh Van Duong    Karam Park    Pilkyu Park

Visual Technology Team, Samsung Research

Seoul, Republic of Korea

{sw123.baek, vinhyd.duong, ka_ram.park, pk79.park}@samsung.com

## Abstract

*This paper introduces a novel multi frame super-resolution network (MFSR) for burst hexadeca Bayer pattern Contact Image Sensor (CIS) images, which includes demosaicing, denoising, multi-frame fusion, and super-resolution. Designing a high-quality reconstruction network poses several challenges as follows: 1) Unlike the Bayer color filter array (CFA) pattern, it is hard to interpolate hexadeca-Bayer pattern since the pixel distance between the same color groups increases; 2) Due to large object motion and camera movements, the final fusion result usually suffers the misalignment resulting a blurry image or ghosting artifacts; 3) The proposed network should be fast and efficient enough to operate in real-time on mobile devices. To overcome these challenges, we propose a novel network, called LatentBurst, which contains: 1) a pyramid align and fusion approach in latent feature to deal with large motion scenario; 2) an efficient UNet-based structure which can run efficiently on mobile device; 3) fine-tuned optical flow estimation and two-step knowledge distillation to reduce domain-gap more effectively. Experimental results in various scenarios demonstrate the effectiveness of our proposed method compared with other state-of-the-art methods.*

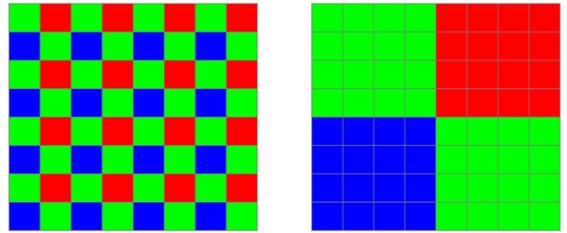

(a) Bayer CFA and hexadeca Bayer CFA

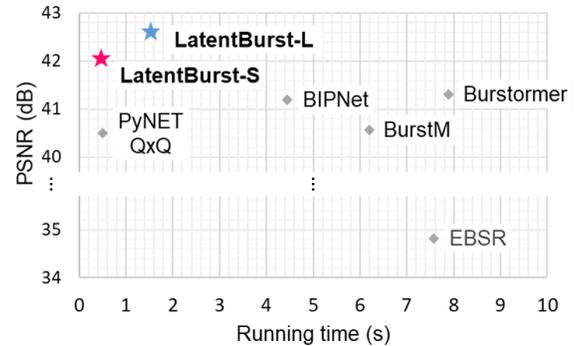

(b) Quantitative comparison with other SOTA methods

Figure 1. (a) Bayer CFA and hexadeca Bayer CFA. Note that the distance between pixels of the same color group in a hexadeca Bayer CFA is greater than that in a Bayer CFA (b) Performance comparison of SR (×2) on burst hexadeca Bayer images (upper-left position indicates better performance). The proposed networks outperform other SOTA multi-frame SR and hexadeca Bayer demosaicing methods, while LatentBurst-S is capable of running on mobile devices.

## 1. Introduction

A digital camera employs its own Image Signal Processor (ISP) to generate an RGB image with a unique visual style from the raw image signal. Most commercial camera ISPs employ handcrafted model-based pipelines for their interpretability and controllability, but these require manual and extensive fine-tuning. In recent years, end-to-end deep learning-based ISPs have drawn increasing attention by surpassing model-based ISP methods [1], offering superior performance without the need for manual adjustments.

 As deep learning-based ISP have advanced and the demand for high-quality smartphone photography has grown, compact camera sensors with smaller pixel sizes have correspondingly evolved. While this miniaturization enables more sophisticated camera systems, it comes at the cost of reduced image quality, particularly in low-light conditions. To mitigate this issue, pixel-binning techniques have emerged, combining adjacent pixels to create larger effective pixel sizes. This approach enhances the signal-to-noise ratio (SNR) in low-light environments while preserving ultra high-resolution capabilities in well-light environments. Regarding this approach, hexadeca Bayer CFA (Figure 1(a)) has been introduced [2], implementing

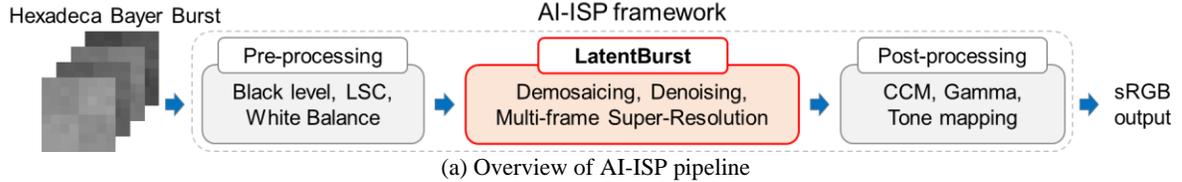

(a) Overview of AI-ISP pipeline

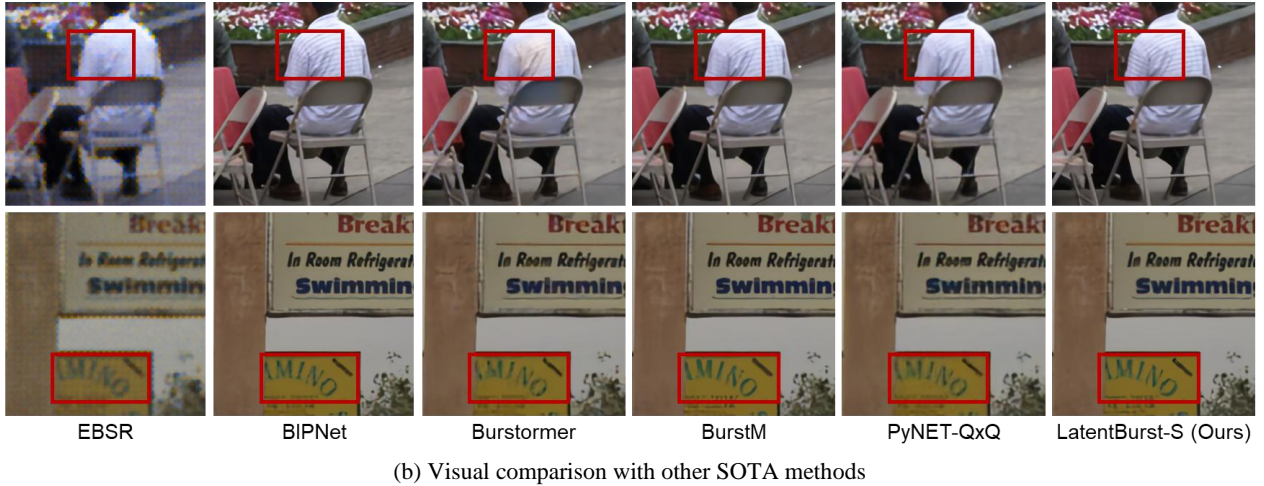

(b) Visual comparison with other SOTA methods

Figure 2. (a) Illustration of LatentBurst in AI-based ISP framework. (b) Visual comparison of all methods for ×2 super-resolution on burst hexadeca Bayer CFA images.

a 4×4 pixel grouping pattern in flagship smartphones. The resolution of these CFA images is approximately up to 16 times greater than that of conventional Bayer pattern CIS images used by many smartphone manufacturers [3, 4]. However, while deep learning-based ISP methods have significantly advanced image reconstruction quality, most existing approaches concentrate on Bayer CFA [5-10], with limited attention given to non-Bayer variants [11-13]. The distinct image statistics produced by non-Bayer CFAs need optimized ISP, particularly for emerging hexadeca Bayer CFA configurations.

The ISP consists of many computational photography steps, including black-level subtraction (BLS), auto white balance (AWB), and lens shading correction (LSC) as pre-processing to burst raw images. Following this, the post-processing stage applies operations such as color correction matrix (CCM), gamma correction and the tone mapping that convert high dynamic range RGB data into 8-bit low dynamic range for final RGB image. Since neural networks can model complex distribution of the given data (i.e., raw images in our case), they easily enable to map the degraded raw input into the high-quality linear RGB (or sRGB) image output. This motivated us to develop a specialized ISP for hexadeca Bayer CFA images on mobile device, where the multi-frame fusion process including super-resolution is replaced with our proposed network as shown in Figure 2(a).

**Problems**. There are existing methods for processing the raw burst images [5-9]. However, none of them can be applied for the ISP framework using non-Bayer CFA for several reasons: 1) The previous methods are designed for Bayer raw burst images, making them unsuitable for processing non-Bayer pattern inputs that exhibit distinct characteristics. Such differences can lead to artifacts or low-quality results. 2) Conventional methods do not take into account the constraints of edge devices, such as memory usage and latency, hindering their deployment on mobile platforms. 3) Compared to Bayer CFA images, large motions are more frequently captured at the pixel level (see Figure 3). Accordingly, addressing these challenges is essential to develop effective and specialized solutions for hexadeca Bayer inputs.

**Contributions**. To solve these problems, we design a new multi frame super-resolution network architecture called LatentBurst. In short, the main contributions of our paper are summarized as follows:

- To our best knowledge, LatentBurst is the first multi frame super-resolution network that transforms burst hexadeca Bayer CFA images into high-quality of single fusion RGB image, while being able to operate in real-time on mobile devices.
- Our proposed network enables to handle complex motion through a pyramid alignment and fusion approach in latent features.

- We introduce a new approach that allocates computational resources across different modules and resolutions, which not only accelerates the network but also maintains the quality.
- We explore an unsupervised optical flow estimation and two-step knowledge distillation strategy, which allow for more effective training of the network.

## 2. Related work

**Learned Image Signal Processor.** The first deep learning based learned ISP was proposed by Ignatov *et al.* [1], who demonstrated that models trained on paired mobile and DSLR images outperform handcrafted traditional ISP. After that, they have organized a challenge in [14] to further improve the quality of learned ISP. Recently, Conde *et al.* [15] have held a raw image restoration and super-resolution challenge, where they introduced larger datasets for learned ISP. However, due to misalignment between raw input and ground truth during the capturing processes, existing datasets for learned ISP may introduce unintended artifacts like ghosting and checkerboard effect. This suggests that the high-quality datasets for learned ISP are utmost important, as they can benefit for producing high-quality reconstructed outputs. Thus, it is essential not only to develop a superior model for learned ISP but also to devise a training strategy that addresses domain-mismatch issues in the training datasets. To this end, we employ unsupervised optical flow estimation and two-step knowledge distillation strategy, which together allow more effective training of the network.

**Multi-frame Super-Resolution.** Recently, after a huge success of deep learning based single image super-resolution [16] and learned ISP [1], many end-to-end learning based multi-frame super-resolution (MFSR) have been introduced [5-10], where Bhat *et al.* [5] introduced the first deep learning based burst super-resolution (DBSR) and Kang *et al.* [10] proposed BurstM, achieving the state-of-the-art results on MFSR up-to-now. However, these methods are only applicable for Bayer CFA images, and only a few studies address demosaicing for single non-Bayer CFA images [11-13]. In addition, to achieve well-aligned frames, they employ the deformable convolutional layers or optical flow-based networks (e.g., SpyNet [17]) to accurately warp complementary information from non-reference frames to the reference frame. However, since hexadeca Bayer CFA images have dynamic scenes with larger motion, existing methods still struggle to handle misalignments, resulting in ghosting artifacts. In detail, it is important to align burst hexadeca Bayer images while keeping color fidelity. Our proposed network addresses this issue by leveraging a fine-tuned optical flow estimation network along with a pyramid alignment and fusion approach.

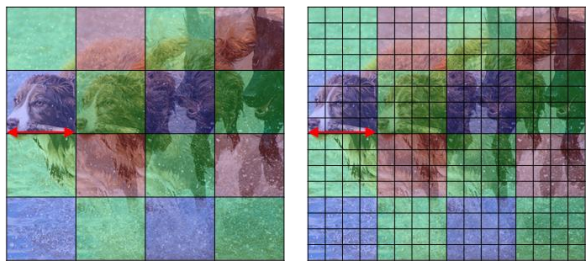

Figure 3. Left: capturing scenario using Bayer CFA with pixel-binning. Right: capturing scenario using hexadeca Bayer CFA. Even under the same capturing scenario, the pixel displacement of fast-moving objects is greater in the hexadeca Bayer pattern, requiring an expanded search range for large motion. For hexadeca Bayer CFA images, the key challenge is to perform demosaicing without introducing false colors while effectively handling large motion.

**Optical Flow Estimation.** Optical flow estimation serves as a fundamental technique for pixel-level motion tracking between video frames, which has found various applications in computer vision tasks, including video enhancement [18, 19] and action recognition [20]. However, these conventional methods often fail to produce accurate results when handling challenging scenarios such as large objects motion. Recent advances in deformable convolutional layer or learning-based methods [21-23] have shown superior performance in both accuracy and efficiency compared to traditional approaches. To predict pixel-wise optical flow for burst hexadeca Bayer CFA, our proposed network used fine-tuned optical flow estimation using unsupervised learning. Thus, the artifacts caused by misalignments between reference frame and other frames were efficiently reduced.

**Knowledge Distillation.** Various knowledge distillation techniques have been developed for different applications. Zagoruyko *et al.* [24] focus on feature distillation methods transferring intermediate feature maps from teacher to student models, while on-line distillation approaches [25-28] simultaneously train both teacher and student models. Adaptive distillation methods [29] enable student networks to learn from multiple teachers. However, most knowledge distillation approaches depend on teacher models pre-trained on paired synthetic datasets. In this case, the student model may become biased toward the teacher model's output including domain gap that occurred in the synthetic datasets, resulting in degradation of image quality. Our approach differs in that student model enables to follow the teacher model, which was distilled from a domain-matched datasets through two-step knowledge distillation.

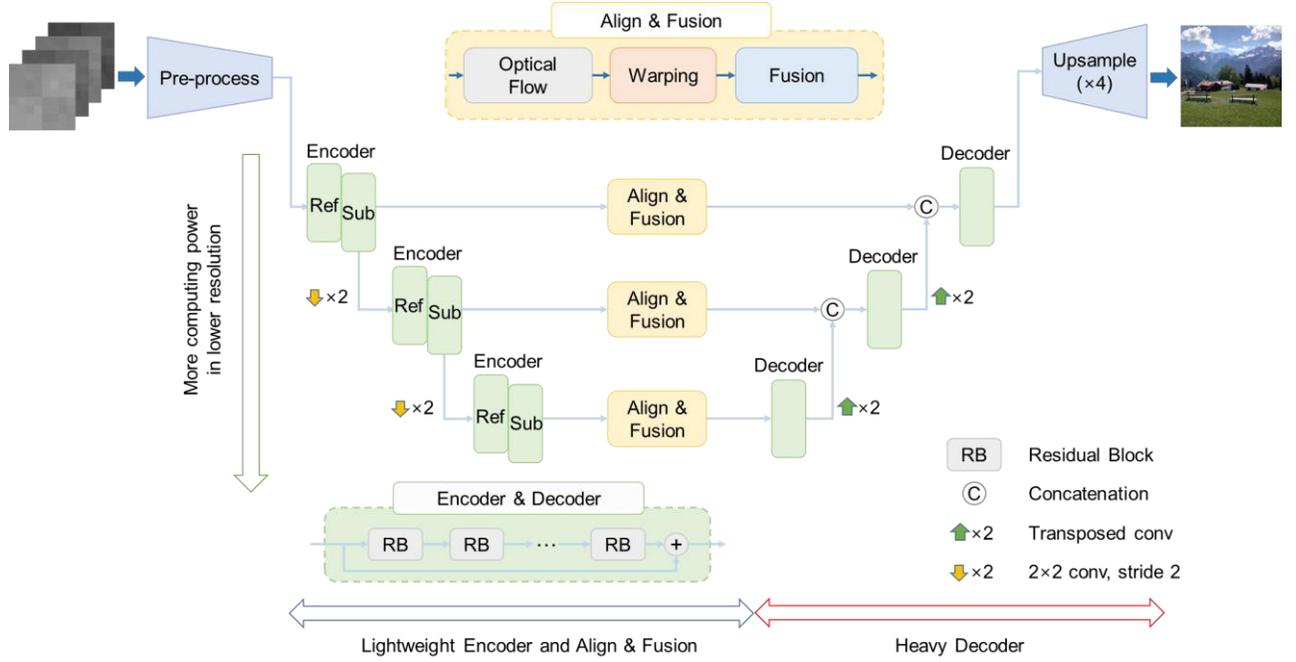

Figure 4. Illustration of the LatentBurst network architecture, where we propose a method to allocate computational resources more efficiently for mobile devices. Moreover, our method incorporates a pyramid alignment and fusion approach that efficiently deals with large motion in burst hexadeca Bayer capture scenario. Note that, the flow estimation performed using SpyNet [17] is omitted from this figure for better visualization.

## 3. Proposed Network

### 3.1. Network architecture

Given a raw burst sequence $\{b_i\}_i^N$, the goal is to fuse these multi frame into a high-quality single fusion frame by taking the shifted complementary information from other non-reference frames into the reference frame. Each image $b_i \in \mathbb{R}^{C \times H \times W}$ is a hexadeca Bayer raw image captured in an image sensor. The proposed network contains mainly five parts: pre-processing, encoder, align and fusion, decoder and upsample parts. The details of each module are described as follows:

**Pre-processing and Encoder.** It should be noted that the hexadeca Bayer pattern exhibits a unique color correlation among different color channels, where each 4×4 block of adjacent pixels shares the same color. Due to this characteristic of high-resolution burst hexadeca Bayer images, it is nearly impossible to directly process the original burst images as input. Therefore, the input images need to be carefully downsampled to preserve their fidelity. To this end, we propose to downsample the original burst images by a factor of four at the very beginning of the network. Specifically, we employ a convolution layer with a kernel size of 4×4 and stride $S = (4, 4)$, which corresponds to each R, G, and B color block in the raw image. The output of this stage is simply denoted as $\widehat{b_i} \in \mathbb{R}^{C \times H/4 \times W/4}$, as follows:

$$\widehat{b_i} = \mathcal{H}_{pre}(b_i) \quad (1)$$

where $\mathcal{H}_{pre}$ denotes the pre-processing layer. After that, we feed the downsampled burst input $b_i^1$ into each encoder block, where each encoder block is composed of multiple residual blocks (RB) (see Figure 4). In particular, the encoder block is divided into two parts: one for reference frame and another for non-reference frames, where they are processed independently. The intermediate feature for each encoder is denoted as $\{f_{enc}^l\}$, as follows:

$$f_{enc}^l = \mathcal{H}_{enc}^l(f_{enc}^{l-1}), l \in [1, L] \quad (2)$$

where $\mathcal{H}_{enc}^l$ denotes the function of each encoder at $l$-th depth, $l \in [1, L]$ is depth level in pyramid structure. After applying each encoder, we employ a convolutional layer with a kernel size of 2×2 and stride $S = (2, 2)$ for downsampling, except for the last depth level. As a result, the feature maps at the deepest level have a spatial resolution of $\mathbb{R}^{C \times H/16 \times W/16}$. It should be noted that we allocate fewer RBs at higher resolutions and more at lower resolutions. Furthermore, we use a smaller number of feature channels than the decoder to reduce the overall memory footprint of network.

**Alignment and Fusion.** To handle a large motion, we propose a multi-scale alignment and fusion approach,

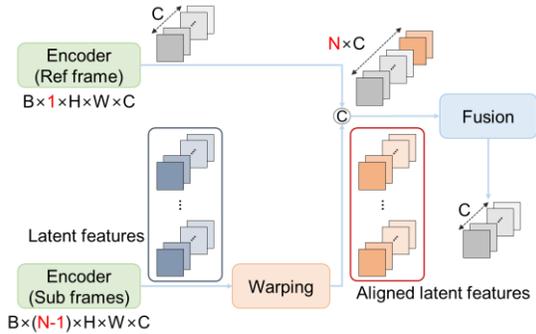

Figure 5. Illustration of align and fusion progress in latent space. The key point is that the alignment and fusion are applied in latent spaces. This approach enables the network to predict and compensate the large object motion within burst hexadeca Bayer frames.

which can effectively correct the misalignment from coarse-to-fine resolution. As illustrated in Figure 5, the core innovation of the proposed network lies in the application of the alignment and fusion process within latent spaces, rather than image spaces. This methodology allows the network to effectively predict and compensate for significant object motion across the burst hexadeca Bayer frames.

In detail, we use SpyNet [17] to estimate the optical flow from raw burst input, where we estimate the back-ward flow from the non-reference burst frames ($\{b_i\}_1^N$) to the reference frame $b_0$ ($N$ denotes number of non-reference frames). Also, let us denote the estimated flows as $\{g_i\}_1^N$. For multi-scale warping and fusion purpose, we need to rescale the estimated flows in advance corresponding to the features in the latent spaces, denoted as $\{f_i\}_1^N$. Then, we warp the feature in the latent spaces from non-reference frames by using the estimated flows as follows:

$$w_i^l = \Im(f_i^l, g_i^l), \ l \in [1, L], i \in [1, N] \quad (3)$$

where $w_i^l$ denotes the aligned latent feature of $i$-th non-reference frames into reference frame at $l$-th depth level by using warping function $\Im(\cdot)$. After that, we concatenate the warped features with those from the reference frame, and then utilize a convolutional layer with kernel size of 1×1 to fuse these features as follows:

$$f_{fusion}^l = \mathcal{H}_{fusion}^l([f_0^l, w_1^l, \ldots, w_N^l]), l \in [1, L] \quad (4)$$

where $f_0^l$ is the feature from the reference frame at the $l$-th depth level, which is concatenated with the aligned features from the non-reference frames $w_i^l$. $\mathcal{H}_{fusion}^l$ denotes the function of fusion layer, and $f_{fusion}^l$ represents the fused feature at $l$-th depth level.

**Decoder and Upsampling.** To achieve a natural reconstruction from latent features to HR images, the decoder is designed with higher computational capacity. Specifically, it employs convolutional layers with more channels than the encoder and a larger number of residual blocks. Following previous works [5-13], an upsampling module based on a pixel-shuffle layer is adopted to restore the processed features to the output resolution.

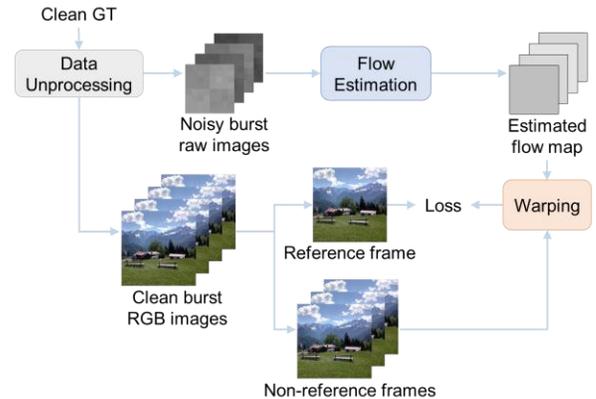

Figure 6. Illustration of fine-tuning optical flow estimation network. Our proposed network employs a fine-tuned optical flow estimation based on unsupervised learning to predict pixel-level motion for burst hexadeca Bayer CFA images. This approach effectively minimizes artifacts resulting from misalignments between the reference frame and other frames in the sequence.

### 3.2. Optical Flow Fine-tuned

Since the use of high-quality flow maps does not generally yield better results in case of multi-frame restoration (burst/video restoration), particularly in noisy and fast-motion scenarios [30], we propose to re-train optical flow estimation network (SpyNet [17] in our case) to better adapt to burst hexadeca Bayer datasets. As shown in Figure 6, we propose a flow estimation fine-tuning strategy using unsupervised learning, where the ground truth flow maps are not available. We utilize the un-processing pipeline [31] to generate paired raw burst input images and corresponding clean images for training the network. This allows the flow estimation network to adapt to the degradation in burst hexadeca Bayer images, and helps the model perform robustly in the presence of noise and large motion. After obtaining flow maps, we warp the non-reference frames to align with the reference frame, and compute the loss accordingly. Note that, warping non-reference frames based on these flow maps enables to help the network only focus on compensating the errors in flow estimation.

### 3.3. Two-Step Knowledge Distillation

Even when synthetic datasets are meticulously generated and the model is well trained on them, a domain gap with real-world captured images may still exist, resulting in low-quality outputs with unintended artifacts. To achieve better performance on real-world hexadeca Bayer images, we apply knowledge distillation framework. Specifically, a large-capacity model with more parameters (LatentBurst-L) is employed as the teacher model, while a lightweight network designed for mobile devices (LatentBurst-S) serves as the student model. Real-world hexadeca Bayer images captured with a 200MP CIS with a hexadeca pattern [2] are used as inputs, and the student model is trained through distillation loss derived from the teacher model's outputs, which serve as a pseudo ground truth.

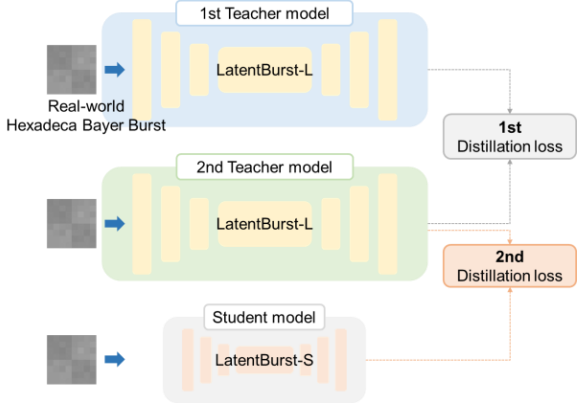

Figure 7. Description of two-step knowledge distillation.

However, the teacher model also suffers from a domain gap since it is trained from the synthetic datasets, which can be simply formulated as follows:

$$x_R \leftrightarrow \hat{x}_k = W_k * I_k + n_k \quad (5)$$

where $x_R$ denotes real-world hexadeca Bayer images, and $\hat{x}_k$ denotes training dataset synthesized from ground truth $I_k$ using degradation kernel $W_k$ and noise $n_k$. Thus, the student model is trained to minimize the difference from the teacher model's output, where the domain gap already exists, as follows:

$$I_k \leftrightarrow I_{\hat{x}_k, x_R} = F_{\hat{x}_k}(x_R) \quad (6)$$

where $F_{\hat{x}_k}$ denotes the student model trained by using $\hat{x}_k$ and $I_{\hat{x}_k, x_R}$ represents the model output that suffers from the domain gap issue. To overcome this discrepancy, we perform another step of distillation using a second teacher model, which has the same architecture, but different pretrained weights (see Figure 7). The second teacher model is then used to guide the student model. As a result, the student model distilled from the second teacher model can be represented as follows:

$$I_k \cong I_{x_R, x_R} = F_{x_R}(x_R) \quad (7)$$

Since $F_{x_R}$ is trained using using real-world hexadeca Bayer images $x_R$, the reconstructed output exhibits less domain discrepancy compared to the existing distillation approach.

## 4. Experiments

### 4.1. Datasets and Implementation details

**Datasets**. In this paper, we use MIT-Adobe FiveK dataset [32], which contains about 5,000 linear RGB images taken with DSLR cameras. After that, we employ the unprocessing pipeline [31] to generate pair training data, following the previous state-of-the-art methods [5-13]. In detail, we render four burst frames with hexadeca Bayer CFA pattern. In training phase, we randomly add large range of translation motion, simulating the dynamic motion scenes. To make the dataset more similar to real burst hexadeca Bayer CIS images, blur kernels and noises are added. In addition, qualitative comparisons[1] of visual quality are conducted using real-captured hexadeca Bayer CIS images.

**Networks Structure**. We build large and small models to evaluate the performance: LatentBurst-L and LatentBurst-S, respectively. In detail, our LatentBurst-S network has several key parameters: the number of feature channel in encoder and decoder $enc_{dim} = [32, 64, 128]$ and $dec_{dim} = [256, 128, 64]$, the number of RBs in encoder and decoder $enc_{nb} = dec_{nb} = [2, 4, 8]$. Therefore, the maximum depth level in our experiments is 3 ($L = 3$). In contrasts, the large model is much heavier, where the number of RBs is set to $enc_{nb} = dec_{nb} = [4, 8, 16]$ for encoder and decoder. Besides, the number of channels in the layers of encoder and decoder are set to $enc_{dim} = [64, 128, 512]$ and $dec_{dim} = [512, 128, 64]$. The fine-tuned model of SpyNet [17] is used in our experiments.

**Implementation details**. We train our network with an end-to-end manner without pre-trained weights. We apply an AdamW optimizer [33] with $10^{-4}$ learning rate, and train our model using NVIDIA A100 GPUs for about 3-4 days. For training with synthetic data, we apply L1 loss. In knowledge distillation phase, we fine-tune the pre-trained model with combined L1 loss and color preserving loss $L_{\text{CIELAB}}$, as follows:

$$L_{\text{CIELAB}} = \sum_{i=1}^{n} |x_i - y_i| \quad (8)$$

where $x_i$ and $y_i$ represent the output and ground truth images in CIELAB color space [34]. Since the distance between same color pixel group in hexadeca Bayer CFA is greater than standard Bayer CFA, it limits to perform color restoration only in the RGB space of the images. Thus, we compute norm vector in CIELAB color space. In case of real-world datasets, we also apply two-step knowledge distillation using LatentBurst-L as a teacher model, where second teacher model is distillated from pre-trained first teacher model that has different weights. The student model is pre-trained using synthetic datasets, and the teacher model's outputs are used as targets to guide the student model.

---

[1] Real-captured hexadeca Bayer CIS images have no ground truth images, so that only qualitative comparison can be performed.

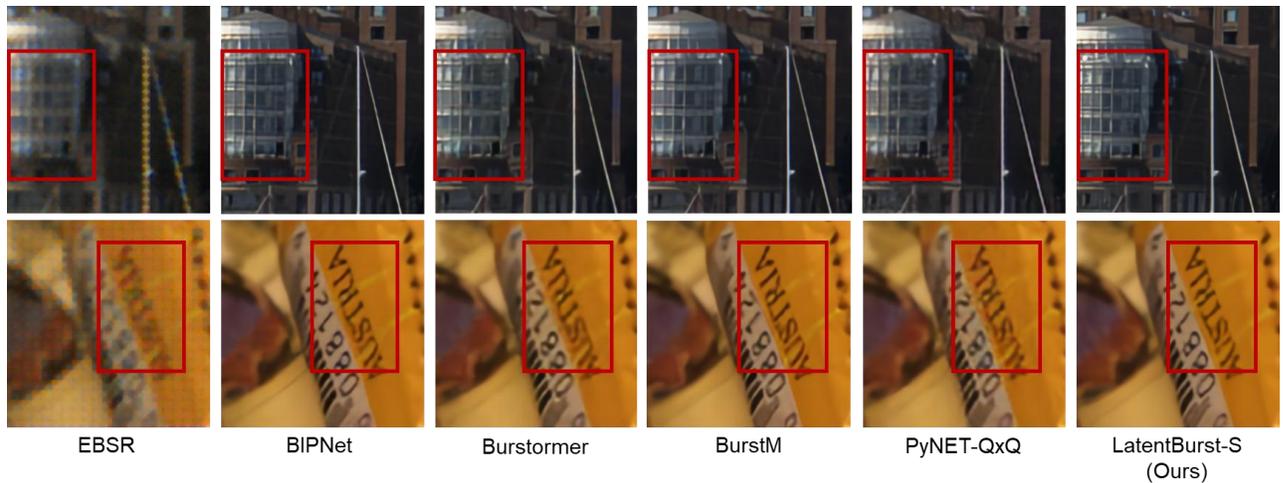

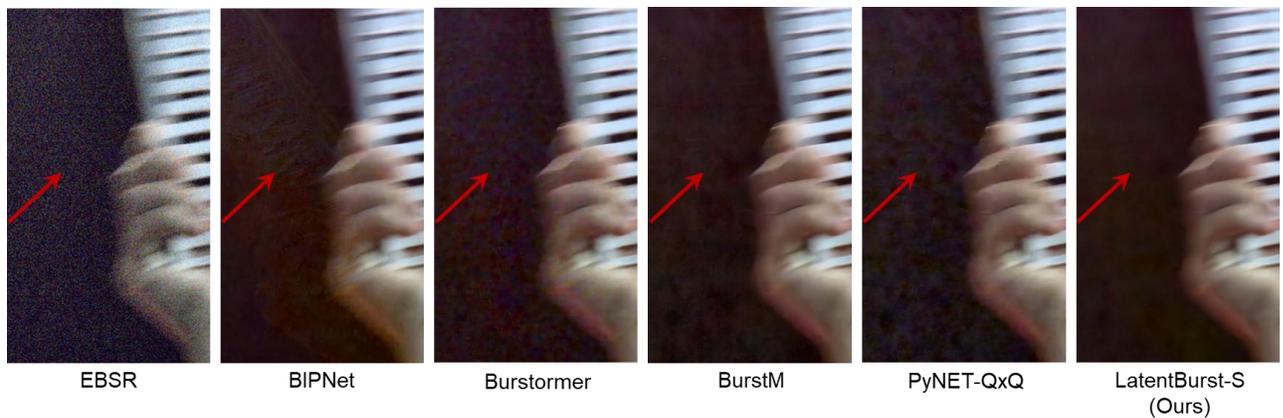

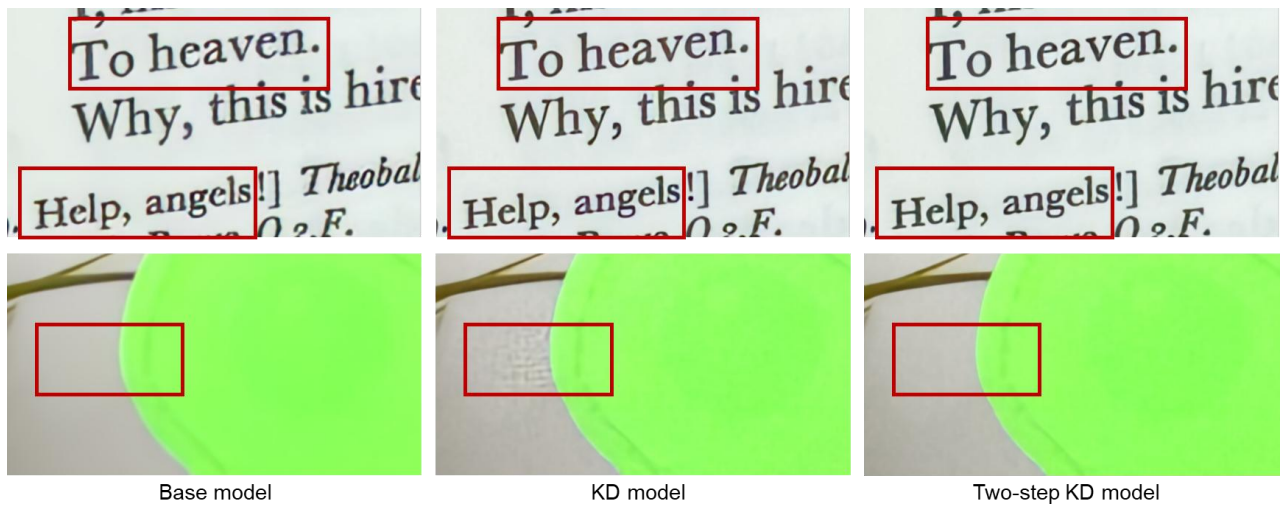

Figure 8: (a) Visual comparison for ×2 super-resolution on synthetic datasets. LatentBurst shows finer details and more accurate colors than the other methods, without introducing artifacts. (b) LatentBurst effectively handles fast-moving objects and performs color interpolation even under low-light conditions. (c) Analysis on two-step knowledge distillation (KD). Compared with existing knowledge distillation method, the proposed two-step knowledge distillation strategy enables the network to remove artifacts more efficiently.

Table 1. Quantitative comparison on synthetic burst hexadeca Bayer images. Red and Blue colors indicate the best and the second performance, respectively.

| Methods | PSNR (dB) | SSIM | Time (sec) | Params. (M) |
|---|---|---|---|---|
| EBSR [6] | 34.82 | 0.9097 | 7.572 | 24.01 |
| BIPNet [8] | 41.19 | **0.9531** | 4.452 | 6.54 |
| Burstormer [9] | 41.31 | 0.9470 | 7.884 | 3.11 |
| BurstM [10] | 40.56 | 0.9498 | 6.195 | 12.58 |
| PyNET-QxQ [13] | 40.49 | 0.9393 | **0.501** | 1.07 |
| LatentBurst-L (ours) | **42.60** | **0.9558** | 1.564 | 1005.38 |
| LatentBurst-S (ours) | **42.03** | 0.9510 | **0.471** | 39.05 |

## 4.2. Experimental Results

We evaluate the performance of our method with several state-of-the-art (SOTA) methods, which are the most up-to-date methods. Since none of existing methods enable to handle burst hexadeca Bayer CFA images directly, several alternative methods can be considered. Thus, quantitative comparisons are performed with SOTA BurstSR [6, 8-10] and single non-Bayer demosaic [13] methods. In case of BurstSR methods, the networks are re-trained according to hexadeca pattern. Also, since PyNET-QxQ [13] handles single hexadeca demosaic without upscaling, we add an upscaling layer based on pixel-shuffle at the final stage and re-train the network. For the fair comparison, all the SOTA methods are trained using the same training procedures as our proposed networks.

**Results on Synthetic Datasets**. Table 1 shows quantitative results of synthetic datasets from MIT-Adobe FiveK compared to other methods. It can be seen that both our proposed models, LatentBurst-L and LatentBurst-S, achieve the best performance in terms of PNSR and SSIM, demonstrating less running time compared to other SOTA methods. In detail, since the most computations are applied in the latent space, the running time of LatentBurst-S can be efficiently reduced despite using more parameters than other methods. Thus, our model is equipped to operate in real-time on mobile devices. Figure 8(a) shows visual comparison with other methods, where we can see that our proposed model can reconstruct better shape for structural information and details. Also, several methods struggled to effectively interpolate missing color between 4×4 pixel grouping patterns. However, our proposed model achieves outstanding results on color interpolation compared to other methods.

**Results on Real-world Datasets**. We evaluate the models on real-world burst hexadeca Bayer CIS images through qualitative comparison. Most conventional methods fail to compensate for large motion (see Figure 8(b)), whereas our proposed model achieves better reconstruction for extremely fast-moving objects without ghosting artifacts. In addition, conventional methods are limited when demosaicing burst raw images captured under low-light conditions due to low SNR, while the proposed model produces better denoised and demosaiced results.

**Effect of Two-Step Knowledge distillation**. As shown in Figure 8(c), the proposed LatentBurst-S demonstrates significant improvements in visual quality for real-world burst hexadeca Bayer images, effectively reducing artifacts. Before applying distillation, several artifacts appeared in the high frequency regions due to the domain gap between the training datasets and the real raw burst datasets. Compared with the existing knowledge distillation methods, the proposed two-step knowledge distillation strategy enables the network to remove artifacts more effectively.

## 5. Conclusion

This paper introduces LatentBurst, a novel multi-frame super-resolution network designed for mobile hexadeca Bayer CIS images, encompassing demosaicing, denoising, multi-frame fusion, and super-resolution. To address the associated challenges, we present a novel network structure comprising: 1) a pyramid alignment and fusion technique applied to latent features to handle large motion scenarios; 2) an efficient UNet-based structure optimized for mobile device performance; and 3) a specialized fine-tuning strategies tailored for hexadeca Bayer CFA patterns; Extensive experimental results across diverse scenarios validate the superiority of our proposed method compared to existing state-of-the-art approaches. Our proposed network demonstrates significant potential to advance deep learning-based ISP, providing critical insights for next-generation high-resolution camera development and mobile implementation.

# References

[1] A. Ignatov, N. Kobyshev, R. Timofte, and K. Vanhoey. DSLR-quality photos on mobile devices with deep convolutional networks. In *ICCV,* 2017.

[2] M. Kwon, I. Ha, Y. Kim, et al. 0.64 $\mu m$ 200 MP Stacked CIS with switchable pixel resolution. In *Proceedings of the International Image Sensor Workshop*, 2023.

[3] A. Kazmi, E. Smith, A. Amer, M. Hafez, and A. Solyman. Comparative image analysis of apple and samsung devices: a technical perspective. In *EICEEAI*, 2023.

[4] J. Joo and H. Alisafaee. Optimization of a mobile phone camera for as-built performance. In *Current Develop-ments in Lens Design and Optical Engineering XXI*, 2020.

[5] G. Bhat, M. Danelljan, F. Yu, L. Van Gool, and R. Timofte. Deep reparametrization of multi-frame super-resolution and denoising. In *ICCV,* 2021.

[6] Z. Luo, L. Yu, X. Mo, Y. Li, L. Jia, H. Fan, J. Sun, and S. Liu. EBSR: feature enhanced burst super-resolution with deformable alignment. In *CVPRW*, 2021.

[7] Z. Luo, Y. Li, S. Cheng, L. Yu, Q. Wu, Z. Wen, H. Fan, J. Sun, and S. Liu. BSRT: Improving burst super-resolution with swin transformer and flow-guided deformable alignment. In *CVPRW*, 2022.

[8] A. Dudhane, S. W. Zamir, S. Khan, F. S. Khan, and M. Yang. Burst image restoration and enhancement. In *CVPR*, 2022.

[9] A. Dudhane, S. W. Zamir, S. Khan, F. S. Khan, and M. Yang. Burstormer: Burst image restoration and enhancement transformer. In *CVPR*, 2023.

[10] E. Kang, B. Lee, S. Im, and K. H. Jin, BurstM: Deep burst multi-scale SR using fourier space with optical flow. In *ECCV*, 2024.

[11] I. Kim, S. Song, S. Chang, S. Lim, and K. Guo, Deep image demosaicing for submicron image sensors. *Journal of Imaging Science and Technology*, 2019.

[12] S. M. A. Sharif, R. A. Naqvi, and M. Biswas, Beyond joint demosaicking and denoising: An image processing pipeline for a pixel-bin image sensor. In *CVPRW*, 2021.

[13] M. Cho, H. Lee, H. Je, K. Kim, D. Ryu, and A. No. Pynet-q×q: an efficient pynet variant for q×q bayer pattern demosaicing in cmos image sensors. *IEEE Access*, 2023.

[14] A. Ignatov, R. Timofte, S. Liu, et al. Learned smart-phone ISP on mobile GPUs with deep learning, mobile AI & AIM 2022 challenge: report. In *ECCV*, 2023.

[15] M. Conde, R. Timofte, Z. Lu, et al. NTIRE 2025 challenge on RAW Image Restoration and Super-Resolution. In *CVPRW*, 2025.

[16] C. Dong, C. C. Loy, K. He, and X. Tang. Image super-resolution using deep convolutional networks. *IEEE TPAMI*, 2016.

[17] A. Ranjan and M. J. Black, Optical flow estimation using a spatial pyramid network. In *CVPR,* 2017.

[18] K. CK Chan, X. Wang, K. Yu, C. Dong, C. C. Loy. Basicvsr: The search for essential components in video super-resolution and beyond. In *CVPR*, 2021.

[19] W. S. Lai, J. B. Huang, N. Ahuja, M. H. Yang. Deep laplacian pyramid networks for fast and accurate super-resolution. In *CVPR*, 2017.

[20] K. Simonyan, A. Zisserman, Two-stream convoluti-onal networks for action recognition in videos. In *NeurIPS*, 2014.

[21] A. Ranjan, M. J. Black. Optical flow estimation using a spatial pyramid network. In *CVPR*, 2017.

[22] D. Sun, X. Yang, M. Y. Liu, J. Kautz. Pwc-net: Cnns for optical flow using pyramid, warping, and cost volume. In *CVPR*, 2018.

[23] Z. Teed and J. Deng. Raft: Recurrent all-pairs field transforms for optical flow. In *ECCV*, 2020

[24] S. Zagoruyko and N. Komodakis, Paying more attention to attention: Improving the performance of convolutional neural networks via attention transfer. In *ICLR*, 2017.

[25] R. Anil, G. Pereyra, A. Passos, R. Ormandi, G. E. Dahl, and G. E. Hinton. Large scale distributed neural network training through online distillation. *arXiv:1804.03235*, 2018.

[26] D. Chen, J. P. Mei, C. Wang, Y. Feng, and C. Chen, Online knowledge distillation with diverse peers. In *AAAI*, 2020.

[27] Q. Guo, X. Wang, Y. Wu, Z. Yu, D. Liang, X. Hu, and P. Luo, Online knowledge distillation via collaborative learning. In *CVPR*, 2020.

[28] I. Chung, S. Park, J. Kim, and N. Kwak, Feature-map-level online adversarial knowledge distillation. In *ICML*, 2020.

[29] S. Du, S. You, X. Li, J. Wu, F. Wang, C. Qian, and C. Zhang, Agree to disagree: Adaptive ensemble knowledge distillation in gradient space. In *NeurIPS*, 2020.

[30] T. Xue, B. Chen, J. Wu, D. Wei, and W. T. Freeman. Video enhancement with task-oriented flow. *Inter-national Journal of Computer Vision*, 2019.

[31] T. Brooks, B. Mildenhall, T. Xue, J. Chen, D. Sharlet, and J. T. Barron, Unprocessing images for learned raw denoising. In *CVPR,* 2019.

[32] V. Bychkovsky, S. Paris, E. Chan, and F. Durand. Learning photographic global tonal adjustment with a database of input/output image pairs. In *CVPR,* 2011.

[33] I. Loshchilov and F. Hutter. Decoupled weight decay regularization. *arXiv:1711.05101*, 2017.

[34] CIE, *Colorimetry*, 3rd edition. CIE Publication No. 15, 2004.